\documentclass[journal]{IEEEtran}
\IEEEoverridecommandlockouts
\usepackage{cite}
\usepackage{xcolor}

\ifCLASSINFOpdf
   \usepackage[pdftex]{graphicx}
\else
  \usepackage[dvips]{graphicx}
\fi
\definecolor{darkgreen}{RGB}{34, 139, 34}
\definecolor{sthlmRed}{RGB}{196,0,100} 
\definecolor{sthlmBlue}{RGB}{0,110,191}

\usepackage[normalem]{ulem} 
\usepackage{soul} 
\usepackage{algorithm,algorithmic}

\usepackage{multirow}
\usepackage{adjustbox}
\usepackage[margin=0.65in]{geometry} 

\newcommand{\transpose}{^\mathsf{T}}

\usepackage{amsfonts}
\usepackage{amssymb}
\usepackage{amsmath}

\begin{document}

\title{Pedestrian Dead Reckoning using Invariant Extended Kalman Filter}

\author{\IEEEauthorblockN{Jingran Zhang, Zhengzhang Yan, Yiming Chen, Zeqiang He and Jiahao Chen*}
\thanks{The authors are with School of Information Science and Technology, ShanghaiTech University, Shanghai 201210, China.\\
*Corresponding author: Jiahao Chen (email: chenjh2@shanghaitech.edu.cn).\\
This work was supported in part by Shanghai Frontiers Science Center of Human-centered Artificial Intelligence, and the MoE Key Lab of Intelligent Perception and Human-Machine Collaboration (ShanghaiTech University).
}
}

\maketitle

\begin{abstract}
This paper presents a cost-effective inertial pedestrian dead reckoning method for the bipedal robot in the GPS-denied environment. 
Each time when the inertial measurement unit (IMU) is on the stance foot, a stationary pseudo-measurement can be executed to provide innovation to the IMU measurement based prediction.
The matrix Lie group based theoretical development of the adopted invariant extended Kalman filter (InEKF) is set forth for tutorial purpose.
Three experiments are conducted to compare between InEKF and standard EKF, including motion capture benchmark experiment, large-scale multi-floor walking experiment, and bipedal robot experiment, as an effort to show our method's feasibility in real-world robot system.  
In addition, a sensitivity analysis is included to show that InEKF is much easier to tune than EKF.
\end{abstract}


\section{Introduction}


Localization constitutes the fundamental capability for autonomous navigation in mobile robotics, serving as the cornerstone for path planning, obstacle avoidance, and environmental interaction \cite{panigrahi2022localization}. 
According to the reliance on external information, robot localization methods can be categorized into two major types: infrastructure-dependent localization and proprioceptive  localization.
Infrastructure-dependent approaches rely on external references such as base stations, global maps, signal sources, or sensor networks for position estimation\cite{azpurua2023survey}. In contrast, proprioceptive localization methods estimate the robot’s states (e.g., position and orientation) solely based on onboard sensory data\cite{roychoudhury2023perception}. Typical sensors used in such systems include inertial measurement units (IMUs)~\cite{hartley2020contact,yang2023multi}, wheel encoders~\cite{zarei2022advancements}, and pressure sensors~\cite{hartley2020contact}.

There are significant differences between indoor and outdoor environments in terms of available external information~\cite{zafari2019survey}. In indoor scenarios, most external positioning sources, such as the global positioning system (GPS) and the global navigation satellite system (GNSS), are generally inaccessible~\cite{ullah2024mobile}. In addition, information obtained from magnetometers is often inaccurate in indoor environments~\cite{markovic2022error}. As a result, positioning in indoor scenarios typically relies solely on proprioceptive sensors, e.g., inertial sensor, for location estimation.
However, the inherent challenge lies in the continuous accumulation of inertial sensor errors during numerical integration of the navigation equations, particularly the unbounded growth of position uncertainty in strap-down inertial navigation systems \cite{wang2019research}.

\begin{figure} 
    \centering 
    \includegraphics[width=1.0\hsize]{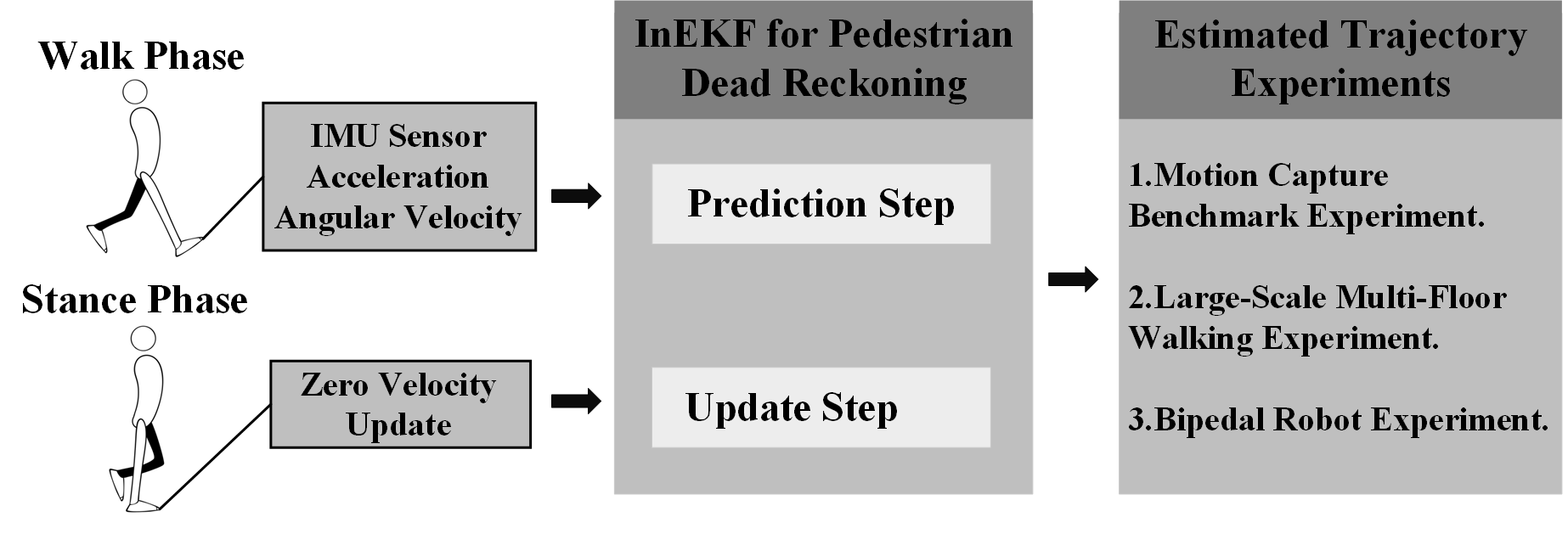} 
    \caption{The pedestrian walking consists of walk phase and stance phase, which correspond to the prediction step and innovation update step in InEKF. 
    } 
    \vspace{-5mm}
    \label{fig:ourwork} 
\end{figure}


During locomotion of humanoid robots and humans, the phase when the foot is in contact with the ground can be regarded as a zero-velocity condition\cite{zhang2021cooperative}. The zero-velocity information from foot contact can be treated as a pseudo-measurement to constrain and correct velocity estimation errors in inertial navigation systems~\cite{hou2020pedestrian}. Therefore, by mounting IMUs on the feet of robots, the zero-velocity information can be effectively utilized to enhance the accuracy of inertial navigation. Hartley \emph{et al.} achieved remarkable results using a fusion strategy based on IMU and contact sensing~\cite{hartley2020contact}. However, their method relies on dedicated contact sensors. Yang \emph{et al.} mounted multiple IMUs on the foot and integrated their measurements within an extended Kalman filter (EKF) framework to realize low‑cost proprioceptive odometry~\cite{yang2023multi}. Nevertheless, the approach employs a relatively basic EKF and does not account for IMU bias.

The integration of zero-velocity updates within EKF has proven effective in significantly reducing drift in foot-mounted inertial navigation systems~\cite{wang2018analytical,fischer2012tutorial}.  
However, conventional EKF implementations rely on Jacobian-based linearization in Euclidean space, which can suffer from inconsistency and degraded performance when confronted with large estimation errors or strongly nonlinear system dynamics.  
To address the limitations, invariant extended Kalman filter (InEKF)~\cite{barrau2016invariant} performs state estimation directly on matrix Lie groups, preserving the underlying geometric structure of the system. This formulation yields trajectory-independent error dynamics under the group-affine condition, leading to improved consistency, more straightforward tuning, and provable stability in deterministic settings~\cite{barrau2018invariant}. Furthermore, the observability properties of InEKF with stationary pseudo-measurements have been analytically characterized~\cite{ramanandan2011inertial}.  
InEKF has been successfully applied to robotic state estimation, encompassing bipedal locomotion~\cite{hartley2020contact,he2024legged,xavier2023multi} and a wide range of navigation and localization tasks~\cite{10318318,potokar2021invariant,guo2023model}. The applications demonstrate the effectiveness of InEKF as an alternative to conventional EKF for accurate and robust proprioceptive odometry.

In this paper, our objective is to achieve low‑cost IMU‑based proprioceptive odometry for bipedal robots without relying on contact sensors. We mount an IMU (that costs about \$1) on the foot rather than on the body,
and then we apply InEKF to reconstruct trajectory of a pedestrian.
The detailed procedure is shown in Fig.~\ref{fig:ourwork}.
It is shown experimentally that the InEKF outperforms EKF and is easier to tune, and when the pedestrian is a ``grumpy'' bipedal robot controlled by a reinforcement learning policy \cite{cuiadapting}, the EKF simply does not work, while InEKF still accurately and stably estimate the system state.
This paper also serves as a self-contained tutorial for applying InEKF to the pedestrian dead reckoning.

\section{Invariant Filtering Basics \\ for IMU based Localization}

The objective is to reproduce the position and orientation of an IMU in the global inertial frame over time against noises. 
The key step in optimal filtering, e.g., Kalman filter, is to derive the Riccati equation that governs how the covariance matrix, denoted by $\Sigma$, propagates \cite{2000-Goodwin.Graebe.ea-book-ControlSystemDesign}.
To derive the Riccati equation, it is required that the state of the filter is governed by linear dynamics.
This is possible if we reformulate the estimation problem using error state (ES) with the aid of matrix Lie group.

\subsection{State Representation using Matrix Lie Group $SE_2(3)$}

Our goal is to reconstruct the orientation $R\in{SO(3)}$ {from the local IMU frame (fixed to pedestrian foot) to the global inertial frame},
as well as the velocity $v\in\mathbb{R}^3$ 
and the position $p\in\mathbb{R}^3$ of the on-board strap-down IMU with respect to the the global inertial frame. 
{These state variables form the square matrix $X{} \in SE_2(3)$ in global inertial frame as}
\begin{equation}\label{eq:Xstate}
X{} \triangleq \begin{bmatrix}
R{} & v{} & p{} \\
0_{1\times3} & 1 & 0 \\
0_{1\times3} & 0 & 1
\end{bmatrix}
\end{equation}
of which the dynamics $f{}(\cdot)$ are \emph{nonlinear} with respect to the yaw, roll, pitch angles because of the use of rotation matrix:
\begin{equation}\label{eq:Xdynamics}
\frac{\mathrm{d}}{\mathrm{d}t} X  =\left[\begin{array}{ccc}
R (\omega)_\times & Ra+g & v \\
0_{1 \times 3} & 0 & 0 \\
0_{1 \times 3} & 0 & 0
\end{array}\right]
 \triangleq f{}(X; \omega,a)
\end{equation}
where $\omega, a \in \mathbb{R}^3$ denote the true angular velocity and linear acceleration measured by an ideal IMU. Since the IMU is mounted on the pedestrian’s foot, both quantities are expressed in the body frame. The notation $(\omega)_{\times}$ represents the $3\times 3$ skew-symmetric matrix associated with $\omega$, and $g \in \mathbb{R}^3$ denotes the gravity vector expressed in the global inertial frame.

\subsection{The Lie Group Error $\eta$ and Error State $\xi$}

In fact, linear dynamics can be derived with only two steps from \eqref{eq:Xdynamics}.
First, we shift our perspective to the estimated state error $\eta\in SE_2(3)$, 
which is essentially a proper choice of \emph{change of state variable}:
\begin{equation}\label{eq:etaDef}
\eta \triangleq \hat{X}{} X{}^{-1}  \in SE_2(3)
\end{equation}
Note the operation between the actual state $X{}$ and the estimated state $\hat{X}{}$ is a ``matrix division'' rather than subtraction,
and in other words, when the estimate is exact, i.e., $\hat X=X$, we have $\eta=I_5$.
One easy way to understand this operation is to regard 
the actual state $X\sim \mathcal{N} \left( \hat{X},\hat{\Sigma} \right)$ as a Gaussian variable on the Lie group, with $\hat X$ as the mean and some proper covariance matrix $\hat{\Sigma}$.

However, it is yet unclear how a covariance matrix $\hat{\Sigma}$ can be defined for an $SE_2(3)$-matrix $X$, and from the definition of $\eta$, one also has
\begin{equation}
\eqref{eq:etaDef}\Rightarrow
X=\hat{X}\eta ^{-1}=\hat{X}\mathrm{e}^{-\xi ^{\land}}
\end{equation}
which implies 
the stochastic behavior of the actual state $X$ with respect to our estimate $\hat X$ stems from a multiplicative noise $\eta^{-1}$~\cite{potokar2024introduction,barrau2018invariant};
and furthermore, $\eta$ can be substituted with a matrix exponential whose exponent $\xi^\land$ belongs to the Lie algebra $\mathfrak{se}_2(3)$:
\begin{equation}
    \xi^{\wedge} = \begin{bmatrix}
    \xi_R \\
    \xi_v \\
    \xi_p
    \end{bmatrix}^\land
    \triangleq \begin{bmatrix}
    (\xi_R)_{\times} & \xi_v & \xi_p \\
    0_{1 \times 3} & 0 & 0 \\
    0_{1 \times 3} & 0 & 0
    \end{bmatrix} 
    \in \mathfrak{se}_2(3) \subset \mathbb{R}^{5\times 5}
\end{equation}
where
the operator $^\land$ sends a $\mathbb{R}^9$ vector into the Lie algebra;\footnote{The exponential map sends a Lie algebra to Lie group:
$\exp(\cdot): \mathfrak{se}_2(3) \to SE_2(3)$.
Vector space can be mapped to the Lie group using 
$\exp (\cdot^\wedge) : \mathbb{R}^{9} \to SE_2(3)$.}
and the three vector components of $\xi$ are marked by the subscripts $_R$, $_v$ and $_p$ corresponding to the error state (ES) variable of orientation, velocity and position, respectively.

Using the error state $\xi$, the covariance matrix $\hat\Sigma$ is now properly defined as in
$\xi  \sim \mathcal{N}(0,\hat\Sigma)$.
In other words, the matrix Lie group theory allows us to properly define the covariance matrix $\hat\Sigma$ regardless of the ``nonlinear state'' $R$.




\subsection{Dynamics of Stochastic Error State $\xi$}

Consider the following naive prediction equation: [cf. \eqref{eq:Xdynamics}]
\begin{equation}\label{eq:XestimateDynamics}
\frac{\mathrm{d}}{\mathrm{d}t}\hat{X}=f{}\left( \hat{X}; \omega, a \right), \quad \hat X|_{t=0}=I_5
\end{equation}
Thus,
the dynamics of error $\eta$ can be derived as follows
\begin{equation}\label{eq:etaDynamics}
\begin{aligned}
&\frac{\mathrm{d}}{\mathrm{d}t}\eta = 
	\frac{\mathrm{d}}{\mathrm{d}t}\left( \hat{X}X^{-1} \right) =\hat{X}\frac{\mathrm{d}}{\mathrm{d}t}X^{-1}+(\frac{\mathrm{d}}{\mathrm{d}t}\hat{X})X^{-1}\\
	&=-\hat{X}X^{-1}\left( \frac{\mathrm{d}}{\mathrm{d}t}X \right) X^{-1}+f\left( \hat{X} \right) X^{-1}\\
	&=-\hat{X}X^{-1}f\left( X \right) X^{-1}+f\left( \eta X \right) X^{-1}\\
	&=-\hat{X}X^{-1}f\left( X \right) X^{-1}+\left[ f\left( \eta \right) X+\eta f\left( X \right) -\eta f\left( I_5 \right) X \right] X^{-1}\\
	&=f\left( \eta \right) -\eta f\left( I_5 \right) 
    \triangleq
    f_\eta\left( \eta \right) 
\end{aligned}
\end{equation}
where the fact that the function $f(\cdot)$ is group affine \cite{potokar2024introduction} has been facilitated to derive the dynamics $f_\eta(\cdot)$ that do not dependent on 
the estimated states $\hat X$ nor the actual state $X$. This is deemed as the key advantage of InEKF over EKF in which the linearization is performed about the estimated states.

Dynamics \eqref{eq:etaDynamics} are still nonlinear with respect to rotations, such that the dynamics of covariance $\hat\Sigma$ cannot be derived yet,
but if we further plug in the first-order approximated relation $\eta =\mathrm{e}^{\xi ^{\land}}\approx I_5+\xi^\land$, 
the linearized dynamics of $\xi$ can further be derived from \eqref{eq:etaDynamics} as follows\footnote{The error state dynamics substituted with the exact relation $\eta =\mathrm{e}^{\xi ^{\land}}$ are
\begin{equation}\label{eq:etaDynamics2}
    \frac{\mathrm{d}}{\mathrm{d}t}\left( \mathrm{e}^{\xi ^{\land}} \right) =f_\eta\left( \mathrm{e}^{\xi ^{\land}} \right) =\left( A_{\xi}\xi \right) ^{\land}+\mathcal{O} \left( \left\| \xi \right\| ^2 \right)     
\end{equation}
In fact, there is a theorem that states the first-order approximation does not cause any error in the estimation for $\eta$ \cite[Theorem~3]{potokar2024introduction}.
}
\begin{equation}\label{eq:Linearization}
\begin{aligned}
f_\eta\left( \eta \right) \approx
   \frac{\mathrm{d}}{\mathrm{d}t}\left( I_5+\xi ^{\land} \right)  &= f{}\left( I_5+\xi ^{\land} \right) -\left( I_5+\xi ^{\land} \right) f{}\left( I_5 \right) 
\\
   &\triangleq \left( A_{\xi}\xi \right) ^{\land}
\end{aligned}
\end{equation}
Note the above linearization process is not dependent on $\hat X$.

We are interested in $\mathbb{R}^9$-vector $\xi$ because it contains all information about the error states $\xi_R,~\xi_v,~\xi_p$ and it is governed by linear dynamics:
\begin{equation}\label{eq:xiDyanmics}
\eqref{eq:Linearization}\Rightarrow
   \frac{\mathrm{d}}{\mathrm{d}t}\xi = A_{\xi}\xi 
   \quad \text{with}~ A_{\xi}=
\begin{bmatrix}
0_{3} & 0_{3} & 0_{3} \\
(g) _\times & 0_{3} & 0_{3} \\
0_{3} & I_{3} & 0_{3}
\end{bmatrix}
\end{equation}
Since the error state $\xi$ evolves along the linear system \eqref{eq:xiDyanmics}, the analytical solution of $\xi(t)$ can be written in terms of transition matrix given its initial value $\xi(0)$ following textbook of modern control theory \cite{2000-Goodwin.Graebe.ea-book-ControlSystemDesign}.
The covariance $\hat\Sigma$ by definition can be calculated as $\hat\Sigma(t)=\mathbb{E}\left[(\xi(t)-0) (\xi(t)-0) \transpose\right]$,
and differentiating $\hat\Sigma(t)$ yields its dynamics (i.e., Riccati equation) 
\begin{equation}\label{eq:SigmaUpdate}
\frac{\mathrm{d}}{\mathrm{d}t}{\hat\Sigma}=A_{\xi}{\hat\Sigma}+{\hat\Sigma} A_{\xi}\transpose
\end{equation}
which relates to an autonomous system, implying that the naive estimator \eqref{eq:XestimateDynamics} is not contaminated by any process noise.


In practice, the IMU measurement noise (denoted by $w^\omega,~w^a$) must be further accounted for, such that the predictor in \eqref{eq:XestimateDynamics} in stochastic settings should be rewritten as
\begin{equation}\label{eq:XestimateDynamicsWithNoise}
\begin{aligned}
\frac{\mathrm{d}}{\mathrm{d}t}\hat{X}&=f{}\left( \hat{X}; \omega+w^\omega, a+w^a \right), \quad \hat X|_{t=0}=I_5
\\
\frac{\mathrm{d}}{\mathrm{d}t}\hat{\Sigma}&=A_{\xi}\hat{\Sigma}+\hat{\Sigma} A_{\xi}\transpose
+\mathrm{Ad}_{\hat {X}}Q\mathrm{Ad}_{\hat {X}}\transpose
\end{aligned}
\end{equation}
where the covariance $\hat{\Sigma}$ 
is driven by the \emph{process noise covariance} $ Q = \mathrm{block\_diag}(\sigma^{\omega}, \sigma^{a}, 0_{3})$,
with $\sigma^\omega$ and $\sigma^a$ respectively the covariance of noises of angular velocity and acceleration measurements, which need to be tuned; 
and finally the adjoint matrix is defined as
\begin{equation}
\text{Ad}_{\hat {X}{}} = \begin{bmatrix}
\hat R{} & 0_{3} & 0_{3} \\
(\hat v{}) _\times \hat R{} & \hat R{} & 0_{3} \\
(\hat p{}) _\times \hat R{} & 0_{3} & \hat R{}
\end{bmatrix}
\label{eq:AdXt}
\end{equation}
Since update of covariance matrix $\hat{\Sigma}$ in \eqref{eq:XestimateDynamicsWithNoise} does not depend on estimate $\hat X$,
the inconsistency problem on calculation of covariance matrix $\hat\Sigma$ faced by EKF \cite{guo2023model} does not exist.







\section{Pedestrian Dead Reckoning with InEKF}


The naive prediction equation in \eqref{eq:XestimateDynamics} needs innovation information on the actual states $R$, $v$ or $p$ for correction, these information is not provided by the IMU.
By utilizing the fact that an IMU (MPU6050) is attached to the foot of a pedestrian (either human or bipedal robot), as shown in Fig.~\ref{fig:experiment_device}.
When the IMU is on the stance foot, 
a stationary pseudo-measurement $v=0$ can be used as the innovation information needed in Kalman filter \cite{ramanandan2011inertial}.

Furthermore, to account for the influence of time-varying measurement biases of IMU 
the angular velocity bias $b^\omega$ and acceleration bias $b^a$ are estimated along with the state $X$ \cite{potokar2021invariant,hartley2020contact}.
It was confirmed in our experiment that bias estimation helps to reduce the drift 
\emph{(e.g., a constant value in the angular velocity measurement)}, especially in the yaw-axis in IMU-based navigation.
However, it is worth pointing out that this is only an engineering treatment in which the Euclidean states $\zeta$ 
and Lie group state $\eta$ are simultaneously estimated \cite{potokar2024introduction}.
\vspace{-2mm}
\subsection{IMU Measurement Model (for State Propagation)}

For low-cost IMUs, biases during the measurement process are often significant and cannot be neglected,  and our adopted IMU measurement model is \cite{potokar2021invariant,hartley2020contact}
\begin{equation}\label{eq:MM}
\begin{aligned}
    \tilde{\omega}{} &= \omega{} + b^{\omega}{} + w^{\omega}{}, & w^{\omega}{} &\sim \mathcal{N}(0, \sigma^{\omega}) \\
    \tilde{a}{} &= a{} + b^{a}{} + w^{a}{}, & w^{a}{} &\sim \mathcal{N}(0, \sigma^{a}) \\
    \frac{\mathrm{d}}{\mathrm{d}t} b^{\omega}{} &= w^{bw}{}, & w^{bw}{} &\sim \mathcal{N}(0, \sigma^{bw}) \\
    \frac{\mathrm{d}}{\mathrm{d}t} b^{a}{} &= w^{ba}{}, & w^{ba}{} &\sim \mathcal{N}(0, \sigma^{ba})
\end{aligned}
\end{equation}
where 
a tilde $\tilde~$ designates actual IMU measurement in practice,
\( w{}^{\omega} \) and \( w{}^{a} \) designate the measurement noise for angular velocity and acceleration, modeled as zero-mean white Gaussian noise with covariances \( \sigma^{\omega} \) and \( \sigma^{a} \), respectively;
\( b^{\omega}{} \) and \( b^{a}{} \) denote the measurement biases of angular velocity and acceleration, respectively, and they are modeled as slowly varying signals driven by Brownian motion, with covariance matrices \( \sigma^{bw} \) and \( \sigma^{ba} \)  governing their stochastic dynamics, respectively. 

With measurement model \eqref{eq:MM}, the naive prediction equation \eqref{eq:XestimateDynamics} is revised with the measured velocity $\tilde\omega$ and acceleration $\tilde a$:
\begin{equation}
\frac{\mathrm{d}}{\mathrm{d}t} \hat{X}  =\left[\begin{array}{ccc}
\hat{R} (\tilde{\omega})_\times & \hat{R} \tilde{a} & \hat{v} \\
0_{1 \times 3} & 0 & 0 \\
0_{1 \times 3} & 0 & 0
\end{array}\right]
-
\hat{X}
\left[\begin{array}{ccc}
(\hat{b}^\omega)^{\wedge} & \hat{b}^a & 0 \\
0_{1 \times 3} & 0 & 0 \\
0_{1 \times 3} & 0 & 0
\end{array}\right] 
\end{equation}
where $\hat{b}^\omega$ and $\hat{b}^a$ denote the estimated angular velocity and acceleration biases.

\subsection{Full Prediction Equation in Discrete-time}

For implementation purpose, 
the discrete-time version of the state propagation is derived using forward Euler method as
\begin{subequations}\label{eq:FullPrediction}
\begin{align}
    \label{eq:prediction:R}\hat{R}_{t+1} &= \hat{R}_t \exp\left( (\tilde{\omega}_t - \hat{b}^{\omega}_t) \Delta t \right) \\
    \label{eq:prediction:v}\hat{v}_{t+1} &= \hat{v}_t + \hat{R}_t (\tilde{a}_t - \hat{b}^a_t) \Delta t + g \Delta t \\
    \label{eq:prediction:p}\hat{p}_{t+1} &= \hat{p}_t +  \hat{v}_t \Delta t + \frac{1}{2} \hat{R}_t (\tilde{a}_t - \hat{b}^a_t) \Delta t^2 + \frac{1}{2} g \Delta t^2 \\
    \label{eq:prediction:bw}\hat{b}^{\omega}_{t+1} &= \hat{b}^{\omega}_t \\
    \label{eq:prediction:ba}\hat{b}^{a}_{t+1} &= \hat{b}^{a}_t
\end{align}
\end{subequations}
where $\Delta t$ is the sampling period, and the subscript $_{t+1}$ suggests the variable is evaluated at the next sampling instant $t+\Delta t$.

Note there is no way we can update the bias estimate during the prediction stage in \eqref{eq:FullPrediction}, and its update would rely on its covariance estimation.
The estimated bias error is directly defined in Euclidean space:
\begin{equation}
\zeta{} = \hat b - b = \begin{bmatrix} \hat{b}^{\omega}{} - b^{\omega}{} \\ \hat{b}^{a}{} - b^{a}{} \end{bmatrix} \in \mathbb{R}^6
\end{equation}
where symbol \( b \) is short-notation:
${b} = 
\begin{bmatrix}
{b^{\omega}}\transpose,~
{b^{a}}\transpose
\end{bmatrix}\transpose
$.
From this point, 
in order for the InEKF implementation to work with the extended error state $[\xi\transpose, \zeta\transpose]\transpose\in \mathbb{R}^{15}$, all vectors and matrices need to be extended.
To this end,
the covariance matrix update equation \eqref{eq:SigmaUpdate} is extended to further include the estimated bias covariance:
\begin{equation}\label{eq:SigmaUpdateExtended}
\hat{\Sigma}_{t+1} = \Phi_t \hat{\Sigma}_t \Phi_t^\top 
+ \Phi_t \,\mathrm{Ad}_{\hat{X}_t,\hat{b}_t} Q \,\mathrm{Ad}_{\hat{X}_t,\hat{b}_t}^\top \Phi_t^\top \Delta t.
\end{equation}
where the process noise covariance matrix \( {Q} \) is\footnote{The components of $Q$ are covariance of the extended noise vector \( w{} \):
$$
w{} = \begin{bmatrix}
w{}^{\omega\transpose}  & w{}^{a\transpose}  & 0_{1 \times 3} & w{}^{bw\transpose} & w{}^{ba\transpose} 
\end{bmatrix}\transpose \in \mathbb{R}^{15}
$$
}
\begin{equation}
{Q} =
\left( \mathrm{block\_diag} \left( \sigma^{\omega},~ \sigma^{a},~ 0_{3\times 3} ,~ \sigma^{b \omega},~ \sigma^{b a}\right) \Delta t \right)^2
\end{equation}
with $\sigma^{b\omega}$ and $\sigma^{ba}$ respectively the covariance of angular velocity bias and acceleration bias to be tuned;
and 
the adjoint matrix and transition matrix are accordingly extended as follows
\begin{align}
\mathrm{Ad}_{\hat{X}_t{},\hat{b}_t{}} &\triangleq \begin{bmatrix}
\mathrm{Ad}_{\hat{X}_t{}} & 0_{9 \times 6}\\
0_{6 \times 9} & I_{6 \times 6}
\end{bmatrix}
\\
\Phi_t 
&= \exp\left(
A_t
\Delta t
\right)
\end{align}
with  \( A_t \) the homogeneous matrix (previously denoted as $A_\xi$) of linear dynamics  extended as~\cite{potokar2021invariant}
\begin{equation}\label{eq:At}
A_t \triangleq \begin{bmatrix}
0_{3} & 0_{3} & 0_{3} & -\hat{R}_t & 0_{3} \\
(g) _\times & 0_{3} & 0_{3} & -(\hat{v}_t) _\times \hat{R}_t & -\hat{R}_t \\
0_{3} & I_{3} & 0_{3} & -(\hat{p}_t) _\times \hat{R}_t & -\hat{R}_t \\
0_{3} & 0_{3} & 0_{3} & 0_{3} & 0_{3} \\
0_{3} & 0_{3} & 0_{3} & 0_{3} & 0_{3}
\end{bmatrix}
\end{equation}

\subsection{Innovation using Stationary Pseudo-Measurement}

\begin{figure}[t] 
    \centering 
    \includegraphics[width=0.7\hsize]{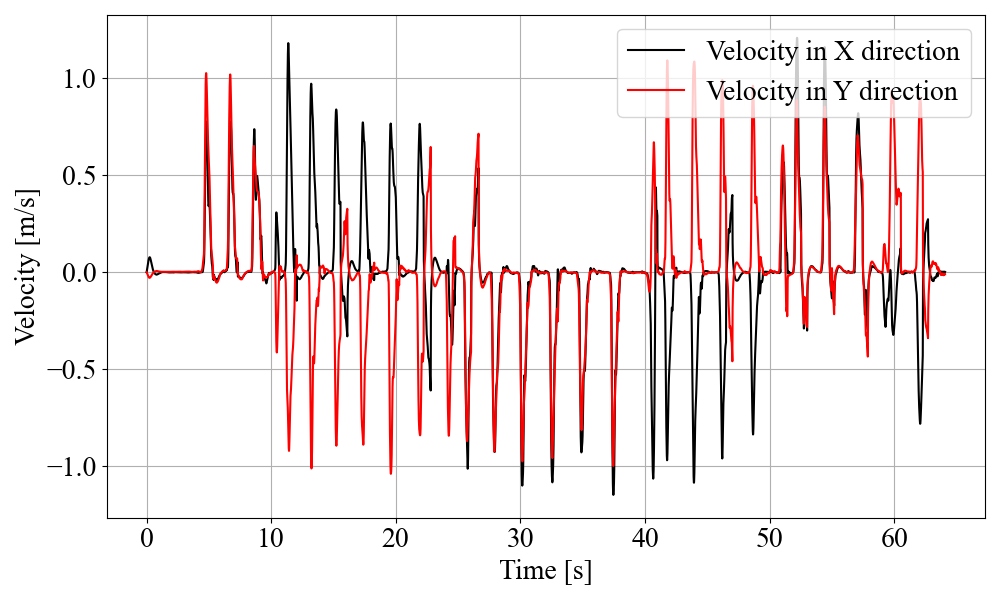} 
    \vspace{-3mm}
    \caption{The estimated velocity in
    pedestrian walking experiment. The velocity is not measured, but apparently there are periodical windows when the velocities drop to zero, which corresponds to the stationary pseudo-measurement.} 
    \vspace{-5mm}
    \label{fig:velocity} 
\end{figure}

The invariant filtering cannot avoid the accumulation of velocity and position errors because there is no measurement with respect to the global inertial frame.
While walking, pedestrian (human or bipedal robots) periodically experiences the stance phase if we assume there is no slipping. 
When IMU is in the stance foot, 
we can make a pseudo measurement of the stationary status of the IMU, that is, all velocities are measure to be zero: 
\( v= \begin{bmatrix} v_x & v_y & v_z \end{bmatrix}\transpose = \begin{bmatrix} 0 & 0 & 0 \end{bmatrix}\transpose \)
which facilitates a zero-velocity update/innovation (ZUPT) \cite{ramanandan2011inertial},
which is confirmed in our experiment, as shown in Fig.~\ref{fig:velocity}.
A threshold-based zero-velocity detection method is employed to identify stationary states. A sliding window is introduced to improve robustness. The detection results activate zero-velocity updates to correct the system state and reduce drift.

In pedestrian dead reckoning, the \emph{invariant measurement} $z$ must be in a proper form that facilitates the calculation of innovation $V$ that is related to the unknown error state $\xi$.
To this end, the following right-invariant measurement model is adopted
for the stationary pseudo-measurement (i.e., $v=0_{3\times1}$) 
\begin{align}
z &= X^{-1}B + \begin{bmatrix} w^z \\ 0_{2\times1} \end{bmatrix} 
  = {\scriptsize\begin{bmatrix}
R\transpose & -R\transpose v & -R\transpose p \\
{0}_{1\times3} & 1 & 0 \\
{0}_{1\times3} & 0 & 1
\end{bmatrix}} B + \begin{bmatrix} w^z \\ 0_{2\times1} \end{bmatrix} 
\nonumber 
\\
&={\scriptsize\left[ \begin{array}{c}
	-R^Tv\\
	-1\\
	0\\
\end{array} \right]}
+\left[ \begin{array}{c}
	w^{z}\\
	0_{2\times 1}\\
\end{array} \right] 
\end{align}
where 
$w^z$ models the slipping between the foot and ground as noise, i.e., $w^z\sim \mathcal{N}(0,M)$ with $M$ the covariance of $w^z$;
vector $B=[0,0,0,-1,0]\transpose$ is introduced here to make sure $z$ only provides the measured velocity,
and it is worth pointing out that invariant measurement $z$ is not zero when the pseudo-measurement $v=0$.
Consequently, the right-invariant innovation $V$ is\footnote{The invariant innovation $V$ is a measurement of the error state $\xi$ because:
    $$
    \begin{aligned}
    	V&=\hat{X}\left( z-\hat{z} \right) =\left( \hat{X}X^{-1}-I \right) B+\hat{X}\left[ \begin{array}{c}
    	w^z\\
    	0_{2\times 1}\\
    \end{array} \right]\\
    	&=\left( \eta -I \right) B+\hat{X}\left[ \begin{array}{c}
    	w^z\\
    	0_{2\times 1}\\
    \end{array} \right] \approx \left( I+\xi ^{\land} \right. -I)B+\hat{X}\left[ \begin{array}{c}
    	w^z\\
    	0_{2\times 1}\\
    \end{array} \right]\\
    	&=\xi ^{\land}B+\hat{X}\left[ \begin{array}{c}
    	w^z\\
    	0_{2\times 1}\\
    \end{array} \right] 
    \end{aligned}
    $$
    }

\begin{equation}
V=\hat{X}\left( z-\hat{z} \right) 
= {\scriptsize\left[ \begin{array}{c}
	\hat{R}R\transpose v-\hat{v}\\
	0\\
	0\\
\end{array} \right]} +\hat{X}\left[ \begin{array}{c}
	w^{z}\\
	0_{2\times 1}\\
\end{array} \right] 
\end{equation}
where \( \hat{z} =\hat{X}^{-1}B \) is the estimated value of the invariant measurement.
That is, an innovation $V$ exists when estimated velocities $\hat v\ne 0$ when IMU is \emph{in stance} such that $v=0$, also known as ZUPT in literature for vehicle navigation \cite{guo2023model}.


To make the invariant innovation $V$ to work with the extended error state vector, 
{the auxiliary matrix} \( \Pi = \begin{bmatrix} I_{3 \times 3} & 0_{3 \times 2} \end{bmatrix} \) is defined such that
$\Pi V = H [\xi\transpose, \zeta\transpose]\transpose$,
with \( H = [0, I, 0, 0, 0] \in \mathbb{R}^{3\times 15} \).
The invariant innovation $V$, which is essentially an error, is also associated with a covariance $S$:
\begin{equation}
S =  H \hat{\Sigma} H\transpose + \hat{R}{} M \hat{R}{}\transpose 
\end{equation}
Following the standard Kalman filter theory, the Kalman gain $K$ is dependent on both state covariance $\hat \Sigma$ and innovation covariance $S$:
\begin{equation}
K = \begin{bmatrix} {K^{\xi}}\transpose & {K^{\zeta} }\transpose \end{bmatrix}\transpose
=
\hat{\Sigma} H\transpose S^{-1} \in\mathbb{R}^{15\times3}
\end{equation}
which is extended to account for measurement bias update, that is,
\( K^{\xi} \in\mathbb{R}^{9\times3} \) denotes the Kalman gain associated with the update of the state estimate \( \hat{X} \), while \( K^{\zeta} \in\mathbb{R}^{6\times3} \) is the Kalman gain used in update the bias estimate $\hat b^\omega, \hat b^a$.

    \begin{algorithm}
    \caption{Discrete-time InEKF for pedestrian dead reckoning}
    \label{alg:inekf_pdr} 
    \begin{algorithmic}[1]
    \STATE $H = \left[ \begin{array}{ccccc} 0, I, 0, 0, 0\end{array} \right];$
    \STATE $\hat{\Sigma}_t = \hat\Sigma_0;$
    \STATE ${Q} =\left( \mathrm{block\_diag} \left( \sigma^{\omega},~ \sigma^{a},~ 0_{3\times3} ,~ \sigma^{b \omega},~ \sigma^{b a}\right) \Delta t \right)^2$ 
    \STATE $[\tilde{a}_x, \tilde{a}_y, \tilde{a}_z] = \tilde{a};~[0,0,g_z]=g;$
    \STATE $\mathrm{ roll }=\arctan (\tilde{a}_y / \tilde{a}_z);$
    \STATE $\mathrm{ pitch }=-\arcsin (\tilde{a}_x / g_z);$ 
    \STATE $\mathrm{ yaw }=0;$
    
    
    
    \STATE $\hat{X}_t=\scriptsize
    \begin{bmatrix}
    \hat{R}_0 & \hat{v}_0 & \hat{p}_0 \\
    0_{1\times3} & I & 0 \\
    0_{1\times3} & 0 & I
    \end{bmatrix};$ \\
    with $\hat R_0=\scriptsize
    \begin{bmatrix}
    \cos (\text{pitch}) & \sin (\text{roll}) \sin (\text{pitch}) & \cos (\text{roll}) \sin (\text{pitch}) \\ 
    0 & \cos (\text{roll}) & -\sin (\text{roll}) \\ 
    -\sin (\text{pitch}) & \sin (\text{roll}) \cos (\text{pitch}) & \cos (\text{roll}) \cos (\text{pitch}) 
    \end{bmatrix},$
    $\hat{p}_0=[0, 0, 0]\transpose$ and $\hat{v}_0=[0, 0, 0]\transpose$.

    \WHILE {receiving data}
        \IF {$\tilde \omega,~ \tilde a$ = IMU measurement}
            \STATE $\hat{R}_{t+1} = \hat{R}_t \exp\left( (\tilde{\omega}_t - \hat{b}^{\omega}_t) \Delta t \right);$
            \STATE $\hat{v}_{t+1} = \hat{v}_t + \hat{R}_t (\tilde{a}_t - \hat{b}^a_t) \Delta t + g \Delta t ;$
            \STATE $\hat{p}_{t+1} = \hat{p}_t + \hat{v}_t \Delta t + \frac{1}{2} \hat{R}_t (\tilde{a}_t - \hat{b}^a_t) \Delta t^2 + \frac{1}{2} g \Delta t^2;$
            \STATE $\hat{b}^{\omega}_{t+1} = \hat{b}^{\omega}_t;$
            \STATE $\hat{b}^{a}_{t+1} = \hat{b}^{a}_t;$
            \STATE $\hat{\Sigma}_{t+1} = \Phi_t \hat{\Sigma}_t \Phi_t\transpose + \Phi_t \mathrm{Ad}_{\hat{X}_t,\hat{b}_t} Q \mathrm{Ad}\transpose_{\hat{X}_t,\hat{b}_t} \Phi_{t}\transpose \Delta t;$ \\
            with $\Phi_t = \exp\left( A_t \Delta t \right)$
            and $A_t$ defined in \eqref{eq:At}.
    
        \ELSIF {$z$ = stationary pseudo-measurement}
            \STATE $V_t = \hat{X_t} \left( z_t - \hat{z}_t \right)$;
            \STATE $S_t^{-1} = (H \hat{\Sigma}_t H\transpose + R_t M_t R_t\transpose)^{-1}$
            \STATE $[K^{\xi}; K^{\zeta}] = K = \hat{\Sigma}_t H\transpose S_t^{-1};$
            \STATE $\hat{X}_t^\dagger = \exp((K^{\xi} \Pi V_t)^\wedge)\hat{X_t};$~ $\hat{X}_t=\hat{X}_t^\dagger$;
            \STATE $\hat{b}_t^\dagger = \hat{b}_t + K^{\zeta} \Pi V_t;$~ $\hat{b}_t=\hat{b}_t^\dagger$;
            \STATE $\hat{\Sigma}_t^\dagger = (I - K H\transpose) \hat{\Sigma}_t;$~ $\hat{\Sigma}_t=\hat{\Sigma}_t^\dagger$;
        \ENDIF
    \ENDWHILE
    \end{algorithmic}
    \end{algorithm}

\begin{figure*}[t] 
    \centering 
    \includegraphics[width=0.9\hsize]{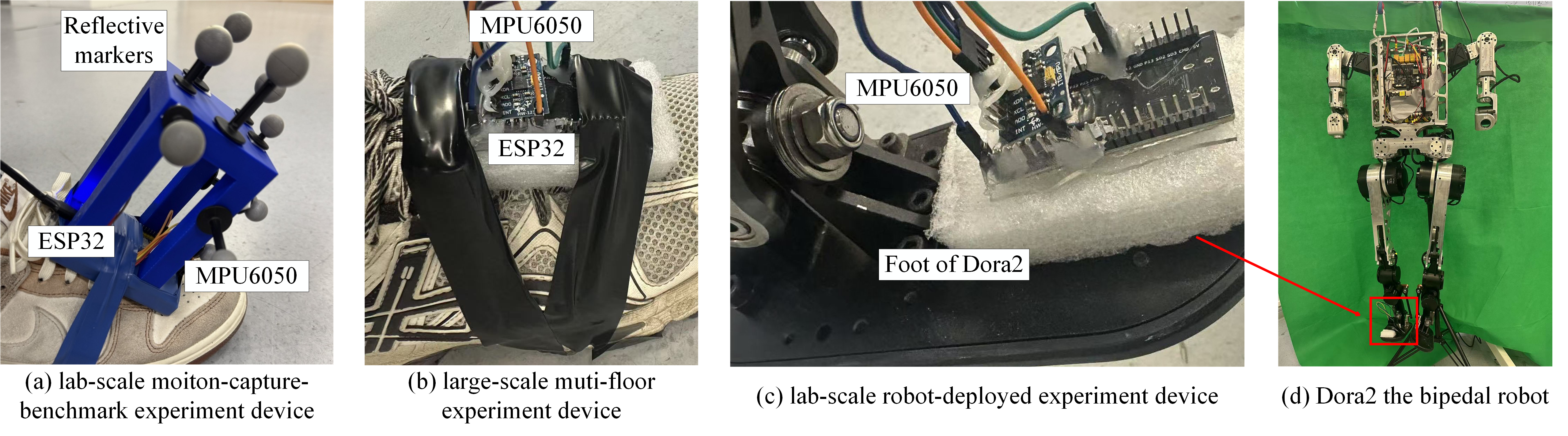} 
        \vspace{-3mm}
    \caption{Photos of the experimental setup. } 
    \label{fig:experiment_device} 
    \vspace{-5mm}
\end{figure*}

Finally, 
the innovation $V$ is used to update estimate $\hat X$ and $\hat b$ via Kalman gain $K$:
\begin{subequations}
\begin{align}
\label{eq:Xinnovated}
\hat{X{}}^\dagger &= \exp((K^\xi \Pi V{})^\wedge)\hat{X{}}
\\
\label{eq:bInnovated}
\hat{b}^\dagger &= \hat{b}{} + K^\zeta \Pi V
\\ 
\hat{\Sigma}^\dagger &= (I-KH) \hat{\Sigma}
\end{align}
\end{subequations}
where a superscript $^\dagger$ indicates the estimate after the stationary pseudo-measurement update,
and auxiliary matrix $\Pi$ is used to make sure $\Pi V\in\mathbb{R}^3$:
\begin{equation}
\begin{aligned}
\Pi V 
    = \Pi \hat{X}\left( z-\hat{z} \right) =\Pi \hat{X}\left( z-\hat{X}^{-1}B \right) =  \Pi \hat{X}z-\Pi B 
\end{aligned}
\end{equation}
with $\Pi B=0$.
Note that \eqref{eq:Xinnovated} is a correction in matrix Lie group, 
and \eqref{eq:bInnovated} is a correction in Euclidean space.
The full procedure of the discrete-time InEKF based pedestrian dead reckoning is listed in Algorithm~\ref{alg:inekf_pdr},
in which the prediction is based on IMU measurement and the innovation is based on the stationary pseudo-measurement.



\section{Experimental Validation Studies}
Fig.~\ref{fig:experiment_device} shows the experimental setups.
A cheap IMU MPU6050 that provides raw measurement $\tilde \omega$ and $\tilde a$,
at a sampling frequency of 100 Hz is adopted.

In traditional pedestrian navigation, the gyroscope's turn rates are commonly used to reliably determine whether the system is in the stance phase \cite{fischer2012tutorial}. 
This paper proposes to determine the stationary status of IMU based on acceleration and angular velocity thresholds.
Looking at the estimated velocities $\hat v$ (because $v$ is not measured) in Fig.~\ref{fig:velocity}, one realizes that a sliding window with a fixed width can be introduced to determine stationary status, in a sense that 
stationary pseudo-measurement $v=0$ happens only when the velocities $\hat v$ at both beginning and end of the sliding window are below the thresholds.
When implemented in robotic experiments, the sliding window width and threshold need to be adjusted accordingly to accommodate the high frequency gait of the bipedal robot.

\subsection{Motion Capture Benchmark Experiment}
\begin{figure}[t] 
    \centering 
    \includegraphics[width=0.45\textwidth]{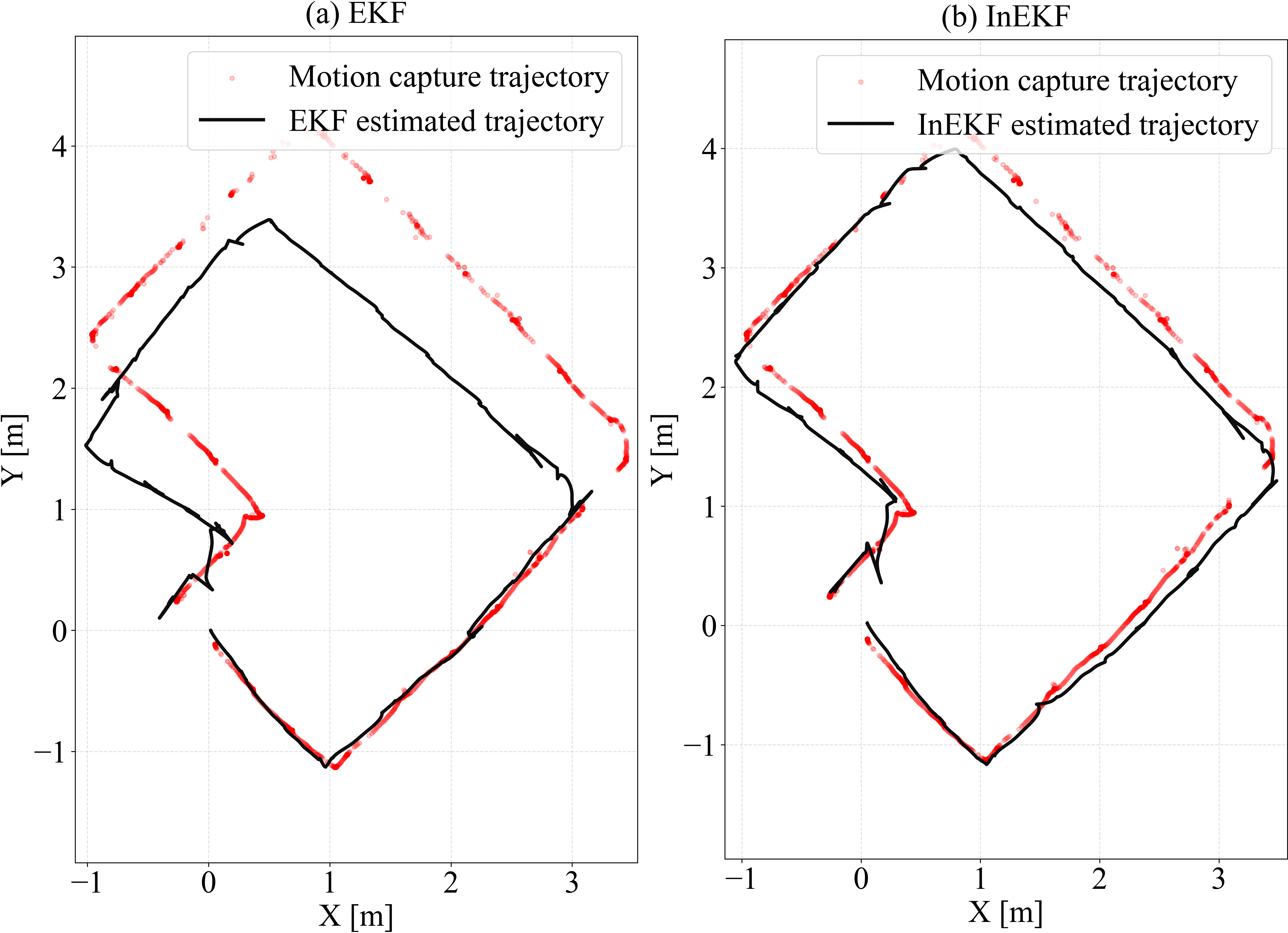} 
        \vspace{-3mm}
    \caption{Comparison to motion capture ground truth: (a) EKF. (b) InEKF.
    }%
    \vspace{-4mm}
    \label{fig:ex1} 
\end{figure}

In the pedestrian dead reckoning problem, acquiring ground truth data for error evaluation is a challenging task \cite{fischer2012tutorial}. Existing research commonly adopts a closed-loop walking route in a pre-investigated map with fixed start and end points to acquire the ground truth, or rely on the GPS data \cite{guo2023model}.
To obtain a high quality motion trajectory for validation purpose, this paper devises a motion capture experiment in a room to collect mocap data as ground truth. 
Fig.~\ref{fig:experiment_device}a illustrates the reflective marker setup for the experiment. 
To ensure that the reflective markers are mounted on a rigid body, a supporting platform was 3D-printed, with MPU6050 mounted on it.

Fig.~\ref{fig:ex1} shows a comparison between the motion capture system's recorded trajectories and those from EKF and InEKF. 
The InEKF-estimated trajectory shows much better alignment with the reference trajectory from the motion capture system.
The reason why the estimated trajectory is not completely aligned with ground-truth, could be due to the fact that the IMU measurement biases $b^\omega,~b^a$ are time-varying. 



\subsection{Large-Scale Multi-Floor Walking Experiment}
\begin{figure}[t] 
    \centering 
    \includegraphics[width=0.46\textwidth]{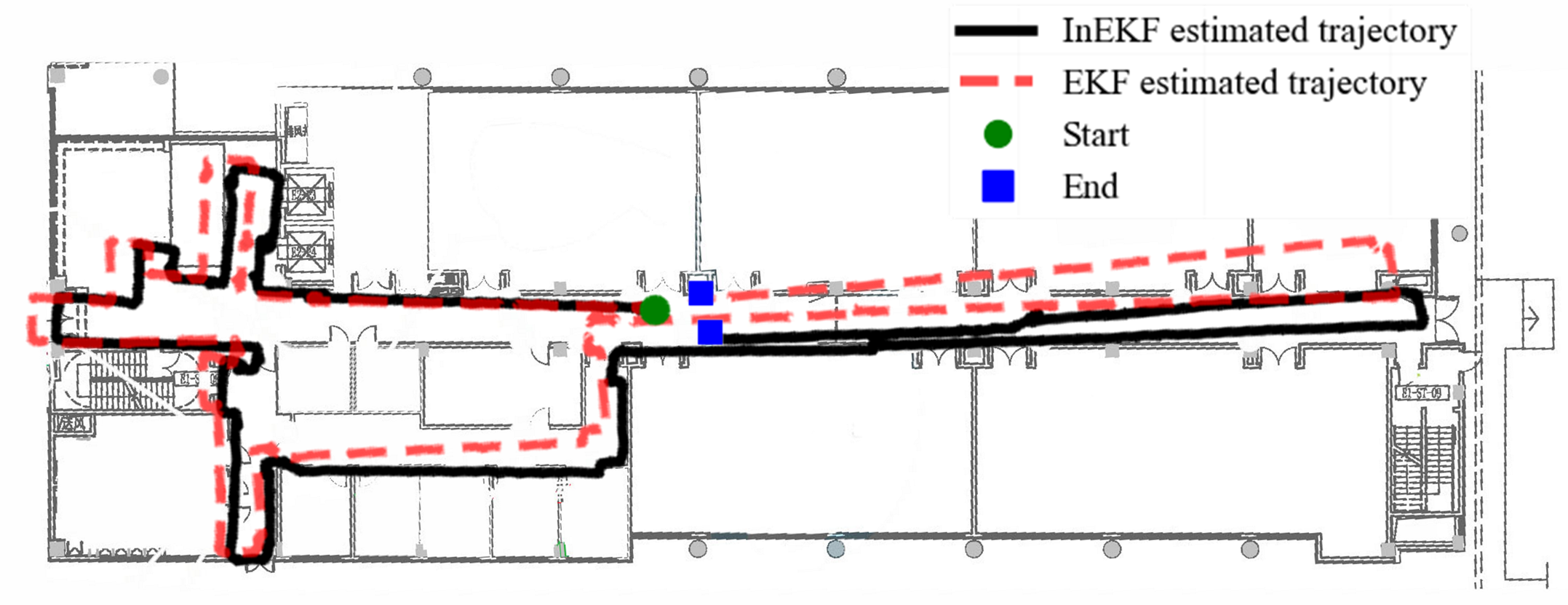} 
    \caption{2D view of indoor walking trajectory upon the map of second-floor of the school at ShanghaiTech University.} %
    \label{fig:ex2} 
\end{figure}

\begin{figure}[t] 
    \centering 
    \vspace{-2mm}
    \includegraphics[width=0.4\textwidth]{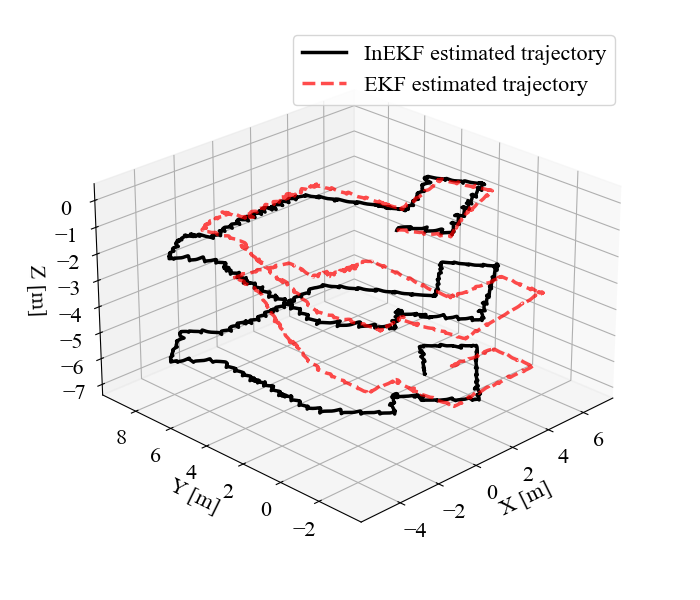} 
    \vspace{-6mm}
    \caption{3D rendering of a multi-floor indoor walking trajectory (from third-floor to first-floor) by stairs. The visualization depicts the vertical and horizontal navigation path across different building levels.} %
    \label{fig:3d} 
    \vspace{-5mm}
\end{figure}

To validate the estimation in large-scale settings,
the pedestrian wore the device as shown in Fig.~\ref{fig:experiment_device}b and performed extensive walking trials.
The pedestrian walked predominantly along a path adjacent to the wall, such that we can use the wall's geometry as a reference trajectory since we don't have motion capture.
Fig.~\ref{fig:ex2} compares the trajectories generated by the EKF and InEKF during an indoor walking experiment in our school. 
The results show that the InEKF trajectory aligns better with the reality, particularly in terms of yaw-axis estimation. 

It is also interesting to see the estimation in the vertical direction.
Therefore, a multi-floor experiment is conducted.
Fig.~\ref{fig:3d} presents a 3D trajectory comparison, including stair-descending motion. 
The EKF method exhibits significant yaw-axis deviation compared to the InEKF. We can learn this provided the fact that the trajectories on different levels should align.

\subsection{Bipedal Robot Experiment}
\begin{figure}[t] 
    \centering 
    \includegraphics[width=0.4\textwidth]{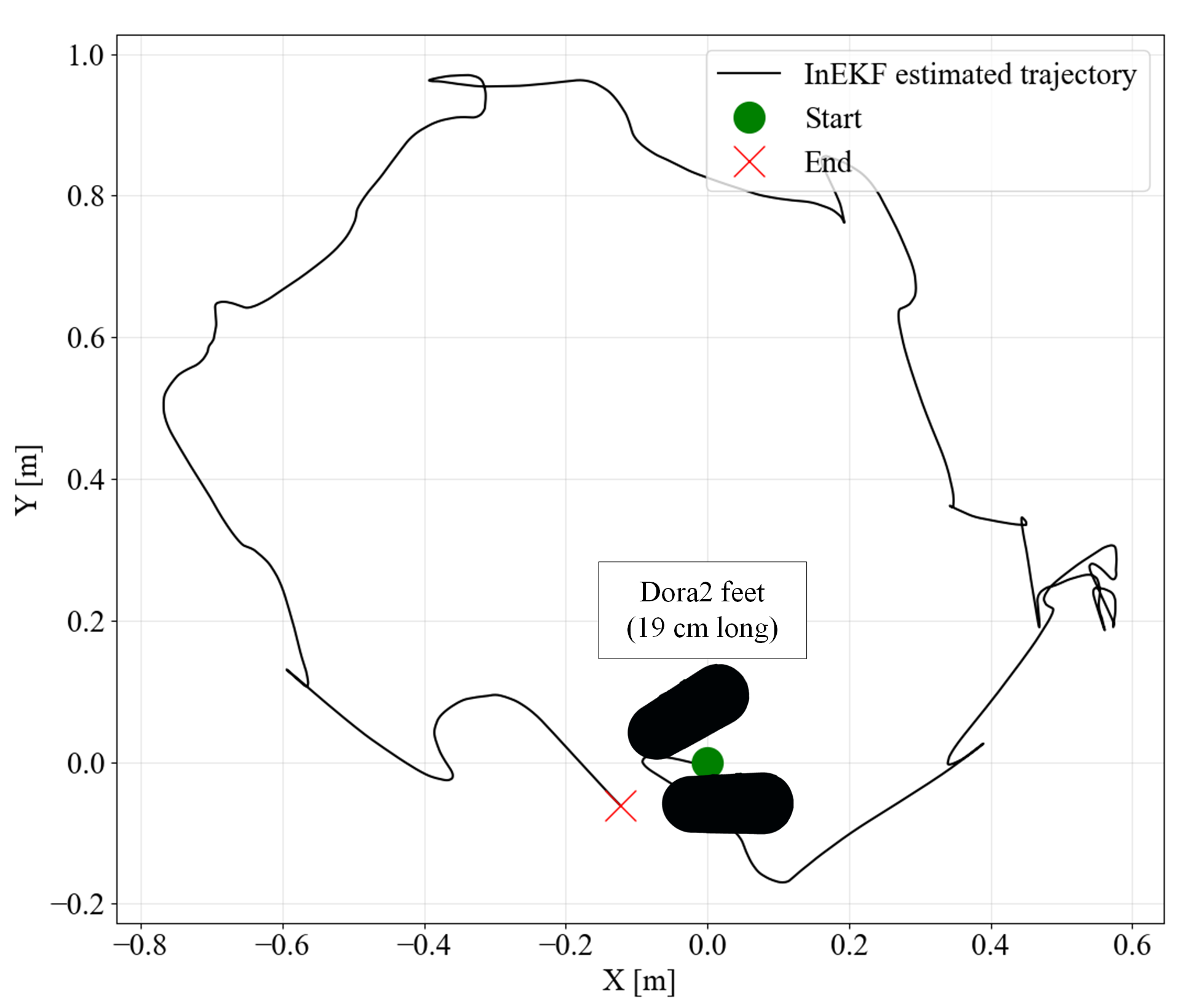} 
    \caption{Walking trajectory of the bipedal robot Dora2 estimated by InEKF. Note the trajectory estimated by EKF is too unsatisfactory to show and is omitted.}
    \vspace{-3mm}
    \label{fig:dora2{}rajectory} 
    \vspace{-3mm}
\end{figure}
The ultimate goal of this study is to apply the proposed method for navigating bipedal robots in GPS-denied environments. 
To assess its feasibility, we devise an experimental setup using Dora2, the bipedal robot, as shown in Fig.~\ref{fig:experiment_device}d. 
As shown in Fig.~\ref{fig:experiment_device}c, the setup consists of an MPU6050 and an ESP32 mounted on the robot's right foot. 
The walking control of the Dora2 robot adopts the policy trained by reinforcement learning elaborated in \cite{cuiadapting}. 
During the experiments, the bipedal robot is controlled by a game-pad to walk along a closed-loop path within the lab.

Fig.~\ref{fig:dora2{}rajectory} illustrates the robot’s movement, with the foot trajectory estimated using the proposed InEKF method. The experimental results show that the InEKF effectively estimates the robot’s foot trajectory and captures the closed-loop characteristic of the path at a gait frequency of approximately 3 Hz, which matches human's normal walking speeds. 
These findings reveal the potential of the proposed method for humanoid robot applications.

\section{Sensitivity Analysis}
\begin{figure}[h!] 
    \centering 
    \includegraphics[width=0.5\textwidth]{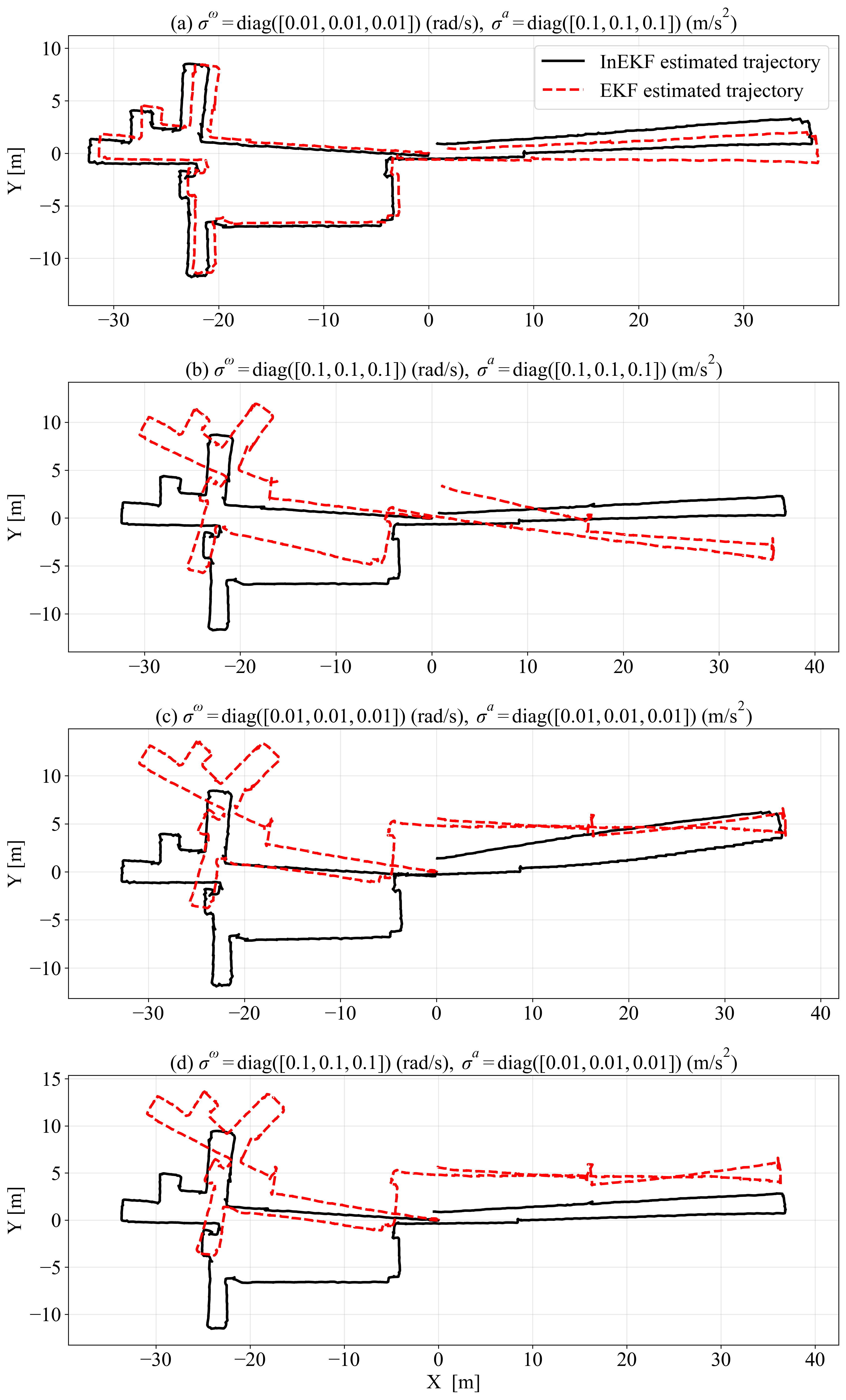} 
    \caption{Comparison of pedestrian dead reckoning trajectory estimates. 
    (a) Both InEKF and EKF give their benchmark performance. 
    EKF is sensitive to value(s) of (b) angular velocity covariance $\sigma^\omega$, (c) acceleration covariance $\sigma^a$, (d) and both covariance.}    
    \label{fig:trajectory_comparison} 
    \vspace{-6mm}
\end{figure}

The traditional EKF strongly depends on carefully tuned process noise covariance $\sigma^\omega,~\sigma^a$ for process and measurement noise, which are typically determined through extensive experimental data or statistical analysis. Variations in these covariance values can significantly degrade EKF performance \cite{wang2017adaptive}. In contrast, InEKF 
exhibits much less sensitivity to process covariance $\sigma^\omega,~\sigma^a$ and is therefore much easier to tune \cite{barrau2018invariant}. This is confirmed with sensitivity analysis experiment shown in Fig.~\ref{fig:trajectory_comparison}.
In Fig.~\ref{fig:trajectory_comparison}a, both methods produce nearly identical trajectories when process noise covariance values are well-calibrated. 
Fig.~\ref{fig:trajectory_comparison}b,~\ref{fig:trajectory_comparison}c and~\ref{fig:trajectory_comparison}d show the trajectory with detuned process noise covariance with a scaling of 10. 
It is observed that EKF trajectory begins to drift significantly,
while the influence to the InEKF trajectory is mild.
One can conclude that InEKF is less sensitive to parameters of process noise covariance \( \sigma^w \) and \( \sigma^a \).

\section{Conclusion}

In this paper, InEKF is applied to the IMU based pedestrian dead reckoning for localization of a robot in GPS-denied scenarios.
Application-specific derivation of the InEKF is set forth, in which the state prediction is based on IMU measurement, while the invariant measurement model is built upon the stationary pseudo-measurement when the IMU is on the stance foot without any mounted contact sensors. Experimental results demonstrate that the InEKF is able to help pedestrians (human and bipedal robot) navigate through indoor environment with acceptable deviations, and outperforms EKF in all aspects, including the sensitivity to process noise covariance.

There exists a non-negligible deviation between the estimated trajectory and the ground-truth trajectory. This deviation is primarily caused by two factors. First, the bias of IMU measurements varies over time, and the conventional bias estimation method defined in Euclidean space is not compatible with the Lie group structure. As a result, this version is referred to as the imperfect InEKF~\cite{potokar2024introduction}, and the convergence of IMU bias estimation cannot be guaranteed.
Second, the current approach relies solely on IMU measurements and zero-velocity pseudo-measurements. Future work may involve developing more accurate contact motion models for bipedal robots to replace the zero-velocity assumption.


\end{document}